\documentclass[runningheads]{llncs}
\usepackage{graphicx}
\usepackage[export]{adjustbox}
\usepackage{algorithm}
\usepackage[noend]{algpseudocode}
\usepackage{amsmath}

\usepackage[utf8]{inputenc} 
\usepackage[T1]{fontenc}    
\usepackage{hyperref}       
\usepackage{url}            
\usepackage{booktabs}       
\usepackage{amsfonts}       
\usepackage{nicefrac}       
\usepackage{microtype}  

\usepackage{colortbl}
\usepackage[dvipsnames]{xcolor}
\newcommand{\mc}[3]{\multicolumn{#1}{#2}{#3}}
\definecolor{Gray}{gray}{0.85}
\definecolor{LightCyan}{rgb}{0.88,1,1}

\newcolumntype{a}{>{\columncolor{Gray}}c}
\newcolumntype{b}{>{\columncolor{white}}c}
\newcommand{\dotr}[1]{#1_{\bullet}} 

\begin{document}
\title{Multi-Task Attention-Based Semi-Supervised Learning for Medical Image Segmentation}
\titlerunning{Multi-Task Attention-Based Semi-Supervised Learning}

\author{
Shuai Chen\inst{1}, Gerda Bortsova\inst{1}, Antonio Garc\'{i}a-Uceda Ju\'{a}rez\inst{1}, Gijs van Tulder\inst{1},  \and
Marleen de Bruijne\inst{1,2}
}

\authorrunning{S. Chen et al.}
%
\institute{
Biomedical Imaging Group Rotterdam, Department of Radiology \& Nuclear Medicine, Erasmus MC, Rotterdam, The Netherlands \and
Department of Computer Science, University of Copenhagen, Copenhagen, Denmark
}
\maketitle  
\begin{abstract}
We propose a novel semi-supervised image segmentation method that simultaneously optimizes a supervised segmentation and an unsupervised reconstruction objectives. The reconstruction objective uses an attention mechanism that separates the reconstruction of image areas corresponding to different classes. The proposed approach was evaluated on two applications: brain tumor and white matter hyperintensities segmentation. Our method, trained on unlabeled and a small number of labeled images, outperformed supervised CNNs trained with the same number of images and CNNs pre-trained on unlabeled data. In ablation experiments, we observed that the proposed attention mechanism substantially improves segmentation performance. We explore two multi-task training strategies: joint training and alternating training. Alternating training requires fewer hyperparameters and achieves a better, more stable performance than joint training. Finally, we analyze the features learned by different methods and find that the attention mechanism helps to learn more discriminative features in the deeper layers of encoders. 


\end{abstract}

\section{Introduction}

\begin{figure}[h]
    \centering
    \includegraphics[width=11cm]{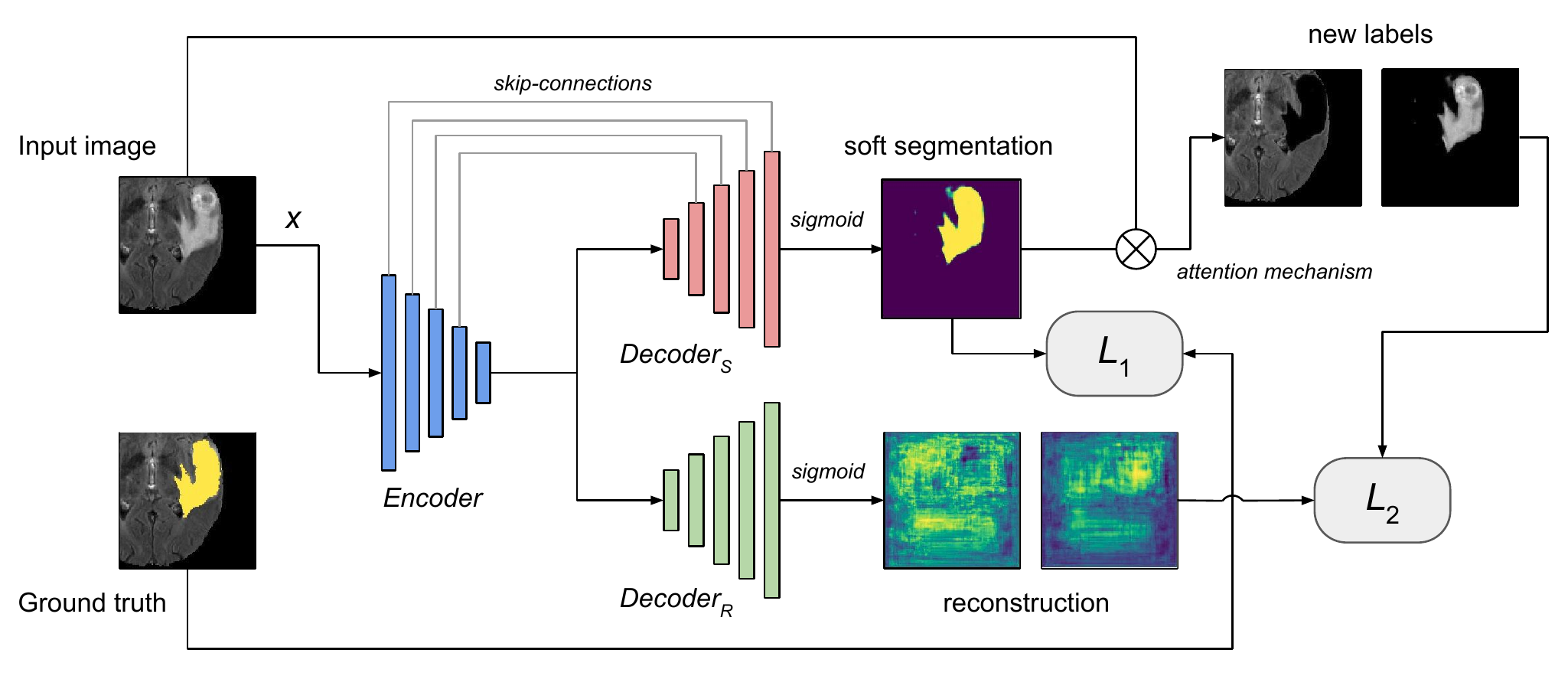}
    \caption{\textbf{Proposed MASSL framework.} Segmentation loss $L_1$ is computed between the soft segmentation prediction and the ground truth. Reconstruction loss $L_2$ is computed between the reconstructed foreground and background prediction and the new labels created using the attention mechanism.}
    \label{MASSL}
\end{figure}

Semi-supervised learning (SSL) uses unlabeled data to improve the generalization performance of a supervised model. This can be useful in medical image segmentation, where manual annotations can be expensive and tedious to produce and are often only available for a small subset of the training data.

One approach to semi-supervised learning is multi-task learning in which the network is trained with an auxiliary objective requiring no manually labeled data, in addition to the target objective using labeled data. This can be done by including an additional autoencoder objective and has been used for image classification (e.g., \cite{Kingma,Rasmus}). Sedai et al. \cite{Sedai} introduced variational autoencoder into semi-supervised segmentation task for the first time, where they train a segmentation autoencoder by learning the encoded embeddings from another pre-trained reconstruction autoencoder and reconstructing the segmentation mask.
However, multi-task learning with image reconstruction is not trivial to combine with popular image segmentation architectures like U-Net \cite{Ronneberger} and its variants \cite{Cicek,Milletari}, which use skip-connections to preserve high-resolution information from their early encoder layers. These skip-connections are not suitable in combination with an autoencoder as the auxiliary task, because they allow the network to copy information from early layers and skip the dimensionality reduction in the autoencoder.

Another semi-supervised approach is creating new pseudo labels for the unlabeled training data, such as self-training \cite{You} and co-training \cite{Xia,Zhou}, to enlist more available training resources. However, the created pseudo labels usually do not have the same quality as the ground truth for the target segmentation objective, which limits their potential for improvements from unlabeled data.

We propose a novel semi-supervised method called Multi-task Attention-based Semi-Supervised Learning (MASSL), in which we combine an autoencoder with a U-Net-like network. Instead of training it to reconstruct the original input \cite{Sedai}, we train the autoencoder to reconstruct synthetic segmentation labels created by the attention mechanism. This encourages our model to learn discriminative features for segmentation from unlabeled images. Although attention is very often applied to supervised learning (e.g., \cite{Schlemper}), to our best knowledge, it has never been combined with semi-supervised learning. Our method has some similarities with self-training \cite{You} and co-training \cite{Xia,Zhou}, which also create new labels for the unlabeled training data on-the-fly. In contrast to these methods, our method creates labels for the reconstruction task. This guides the unsupervised auxiliary task to learn a more discriminative latent representation from unlabeled data than that learned by the traditional reconstruction network, which does not consider class differences.

Our contributions are summarized as follows.
Firstly, we propose a novel multi-task semi-supervised learning method and study its performance in combination with two training strategies.
Secondly, we evaluate our method on two segmentation problems (brain tumors and white matter hyperintensities), demonstrating that it outperforms a fully-supervised CNN baseline, two pre-training approaches, and multi-task learning without the proposed attention mechanism. 
Thirdly, we investigate how the attention mechanism affects the features learned by the encoder and show that it helps the deeper layers to learn more discriminative features.

\section{Methods}

Our semi-supervised learning method is shown in Fig.~\ref{MASSL}.
It consists of a segmentation and a reconstruction networks sharing the same encoder, and an attention mechanism connecting the two tasks.

\subsection{Architecture and Loss Functions}

The segmentation CNN architecture, similarly to U-Net \cite{Ronneberger}, has skip-connections, allowing the transfer of fine details from shallower layers of the encoder to its decoder, and is trained using Dice objective $L_1$ on labeled images only (see Fig.~\ref{MASSL}).
The reconstruction network has a decoder without skip-connections, resulting in an autoencoder, and is trained using mean squared error (MSE) on both labeled and unlabeled images.

In the baseline version of our method, the output of the reconstruction network is optimized to predict the input image.
We call this method Multi-task SSL (MSSL) in the remainder of the paper.

In the attention-based version of our method, which we call Multi-task Attention-based SSL (MASSL), we reconstruct separately background and foreground parts of the image, as defined by the soft predictions $\tilde{y}$ obtained from the segmentation network.
The foreground and background objectives are weighted by the size of the respective segmentation masks:
\begin{equation}
L_2 = \frac{\sum_i{\tilde{y}_b^{(i)}}}{n}MSE[\hat{y}_{b}, {x} \odot \tilde{y}_b] + \frac{\sum_i{\tilde{y}_f^{(i)}}}{n}MSE[\hat{y}_{f}, {x} \odot \tilde{y}_f] 
\end{equation}
where $\dotr{\hat{y}}$ and $\dotr{\tilde{y}}$ are the predictions of the reconstruction and segmentation paths, respectively, for the background ($b$) and foreground ($f$); $n$ is the number of voxels in input image $x$; $\odot$ is element-wise product.
Note that the gradient does not propagate through $\tilde{y}$ to the segmentation decoder.
We hypothesize that infusing reconstruction labels with segmentation predictions will lead to learning better features in the deeper layers of the encoder and hence better segmentation.
The objective terms are weighed to prevent over-emphasizing the importance of foreground reconstruction.

\subsection{Training Strategy}

The two tasks of the MSSL and MASSL networks can be optimized jointly or alternatingly:
\paragraph{Joint training:}Given a minibatch containing an equal number of labeled samples $x_L$ and unlabeled samples $x_U$, the unlabeled samples $x_U$ are first segmented using the most recent segmentation network parameters, to create the foreground and background images for the reconstruction task. Then, the weights of the entire network are updated by optimizing the objective function of both segmentation and reconstruction tasks. The loss is a linear combination of segmentation and reconstruction losses controlled by the hyperparameter $\gamma \in [0, 1]$:
\begin{equation}
\label{joint_loss}
L(x_L, x_U)=\gamma{L_1(x_L)} + (1-\gamma)L_2(x_U)
\end{equation}

\paragraph{Alternating training:}For each epoch, labeled and unlabeled images are randomly sampled by the same amount (the smaller amount of either labeled and unlabeled images) from their corresponding training sets. A minibatch contains either labeled samples $x_L$ or the same amount of unlabeled samples $x_U$. The two types of batch are alternated during training. The weights of the segmentation path and reconstruction path are updated individually according to the given batch type and the corresponding loss:

\begin{equation}
\label{alter_loss}
    L(x)= 
\begin{cases}
    L_1(x),& \text{if } x= x_L\\
    L_2(x),& \text{if } x= x_U
\end{cases}
\end{equation}

\section{Experiments}

\subsubsection{Data}
\label{Data}
We use the public data from the BraTS 2018 Challenge \cite{Menze,Bakas} and the White Matter Hyperintensities 2017 Challenge\footnote{https://wmh.isi.uu.nl/}:

\textit{BraTS18:} 220 MRI scans from patients with high grade glioma are randomly split into 120, 50, 50 scans for training, validation and testing respectively, with 5-fold Monte Carlo cross-validation. To simplify comparison between the different segmentation tasks we perform binary classification and segment only the whole tumor, including all four tumor structures, and use only the FLAIR sequence.

\textit{WMH17:} There are 60 FLAIR MRI scans provided with corresponding manual segmentations of white matter hyperintensities (WMH). The scans are acquired at three sites, 20 at each site. In our experiments, we use 30 scans for training, 10 for validation and 20 for testing, ensuring approximately equal numbers for each site in each of the three sets. We use 5-fold Monte Carlo cross-validation.

\subsubsection{Network and hyperparameters}
The network layout is shown in Fig. \ref{MASSL}. Our network is inspired by the UNet \cite{Ronneberger} architecture but has several differences. The input size of the network is $128\times{128}\times{32}$. There are 5 resolution levels in the encoder and in each of the decoders. Each level consists of two $3\times{3}\times{3}$ convolution layers using zero-padding, instance normalization \cite{Zhou1} and \textit{LeakyReLU} activation functions, except for the last layer of both decoders which use \textit{sigmoid} to make the final prediction. There is an average pooling/upsampling layer between each level. The number of feature channels is 16 in the first level, which is doubled/halved after each pooling/upsampling to a maximum of 256 features at the deepest level. The feature maps in the segmentation upsampling path are concatenated with earlier ones through skip-connections. The reconstruction network has the same architecture as the segmentation network but does not have skip-connections. For joint training, we use one \textit{Adam} optimizer to optimise the loss in Eq.~\ref{joint_loss}. For alternating training, we use two individual \textit{Adam} optimizers to optimize the two types of loss in Eq.~\ref{alter_loss} separately. Based on the performance on the validation sets, we set the initial learning rate to 0.01 and 0.001 for the segmentation and reconstruction tasks respectively. Random rotation, scaling, and horizontal flipping are applied as data augmentation.

\subsubsection{Feature analysis}
We use linear regression analysis to evaluate how well the features can discriminate between foreground and background regions in the last layer of every encoder level. We consider each voxel as an individual sample, using its values in each feature map as the regression variables. The label for each voxel is obtained by the taking binary segmentation ground truth and then down-sampling this with average pooling to the required resolution.

\section{Results}
The segmentation results are shown in Table \ref{BraTS} and Table \ref{WMH}. For the semi-supervised setting (first two colomns), there is no overlap between labeled and unlabeled data. For the fully-supervised setting (last column), all the images are used as labeled and unlabeled data. For \textit{Pretrain(Dec)} we pretrain the reconstruction network with unlabeled data first and then train the decoder path of the segmentation network with labeled data, while keeping the encoder part fixed to ensure that the segmentation task can only use the features learned from unlabeled images. For \textit{Pretrain(CNN)} we pretrain the reconstruction network with unlabeled data first and then train the whole segmentation network using labeled data, which allows the network to fine-tune the encoder parameters if necessary. \textit{MASSL} and \textit{MSSL} are the proposed multi-task SSL methods with and without the attention mechanism, where $\gamma$ and \textit{alter} indicate joint training and alternating training respectively. For joint training, we tried $\gamma=0.5,  0.7, 0.9$ and the network did not converge when $\gamma=0.5$. The results show that MASSL(alter) achieves the best segmentation performance of all methods. The joint training strategy achieved a slightly lower performance than alternating training, which also varied a lot between different labeled/unlabeled data splits, reflecting the instability of the joint training strategy and the difficulty of tuning $\gamma$.

\begin{table}[h]
  \setlength{\tabcolsep}{4pt}
  \caption{\textbf{BraTS18 results.} Dice similarity coefficient, averaged over all cross-validations. The last column uses all labeled images also as unlabeled images, except for the CNN which could only use labeled images. \textcolor{black}{*}: significantly better than CNN (p<0.05). \textsuperscript{\textcolor{black}{$\diamond$}}: significantly worse than MASSL(alter) (p<0.05). P-values are calculated by a two-sided t-test in each column.}
  \label{BraTS}
  \centering
  \begin{tabular}{llll}
    \toprule
    \#Labeled (unlabeled)    & \mc{1}{b}{20~(100)}~ &  \mc{1}{b}{50~(70)}~ &  \mc{1}{b}{120~(120)}~\\
    \midrule    
   CNN & 0.6939$(\pm0.03)$ & 0.7054$(\pm0.03)$& 0.7342$(\pm0.02)$  \\
    Pretrain(Dec) & 0.6948$(\pm0.03)$\textsuperscript{\textcolor{black}{$\diamond$}} & 0.6886$(\pm0.03)$\textsuperscript{\textcolor{black}{$\diamond$}}  & 0.7162$(\pm0.02)$\textsuperscript{\textcolor{black}{$\diamond$}} \\ 
    Pretrain(CNN) & 0.7125$(\pm0.03)$\textsuperscript{\textcolor{black}{$\diamond$}} & 0.7167$(\pm0.03)$\textsuperscript{\textcolor{black}{$\diamond$}}  & 0.7530$(\pm0.02)$\\
    MSSL($\gamma$=0.7) & 0.6140$(\pm0.04)$\textsuperscript{\textcolor{black}{$\diamond$}}  &  0.7433$(\pm0.02)$\textcolor{black}{*}\textsuperscript{\textcolor{black}{$\diamond$}}  &  0.7310$(\pm0.02)$\textsuperscript{\textcolor{black}{$\diamond$}}\\
       
    MSSL($\gamma$=0.9) & 0.6297$(\pm0.03)$\textsuperscript{\textcolor{black}{$\diamond$}}  &  0.7466$(\pm0.02)$\textcolor{black}{*}\textsuperscript{\textcolor{black}{$\diamond$}}  & 0.7568$(\pm0.02)$ \\
   MSSL(alter) & 0.7261$(\pm0.03)$\textcolor{black}{*}\textsuperscript{\textcolor{black}{$\diamond$}} & 0.7462$(\pm0.03)$\textsuperscript{\textcolor{black}{$\diamond$}}  & 0.7461$(\pm0.02)$\\
   \midrule
       MASSL($\gamma$=0.7) & 0.6096$(\pm0.03)$\textsuperscript{\textcolor{black}{$\diamond$}} & 0.7412$(\pm0.02)$\textcolor{black}{*}\textsuperscript{\textcolor{black}{$\diamond$}} & 0.7589$(\pm0.02)$\\
    MASSL($\gamma$=0.9) & 0.6168$(\pm0.04)$\textsuperscript{\textcolor{black}{$\diamond$}} & 0.7159$(\pm0.03)$\textsuperscript{\textcolor{black}{$\diamond$}} & 0.7660$(\pm0.02)$\textcolor{black}{*}\\
   \textbf{MASSL(alter)} & \textbf{0.7553}$(\pm0.03)$\textcolor{black}{*}  & \textbf{0.7710}$(\pm0.02)$\textcolor{black}{*}& \textbf{0.7702}$(\pm0.02)$\textcolor{black}{*}\\ 
 
    \bottomrule
  \end{tabular}
\end{table}

The results of the feature analysis are shown in Table \ref{Reg}. The higher $R^2$ scores indicate that the features learned with MASSL are more  discriminative in the deeper levels than those of CNN and MSSL. This supports our hypothesis that the attention mechanism can make the deeper layers of the encoder learn more discriminative features while still also optimizing the reconstruction objective.

\begin{table}[h]
  \setlength{\tabcolsep}{4pt}
  \caption{\textbf{WMH17 results.} Dice similarity coefficient, averaged over all cross-validation. The last column uses all labeled images also as unlabeled images, except for the CNN which could only use labeled images. \textcolor{black}{*}: significantly better than CNN (p<0.05). \textsuperscript{\textcolor{black}{$\diamond$}}: significantly worse than MASSL(alter) (p<0.05). P-values are calculated by a two-sided t-test in each column.}
  \label{WMH}
  \centering
 \begin{tabular}{llll}
    \toprule
    \#Labeled (unlabeled)    & \mc{1}{b}{10~(20)}~ &  \mc{1}{b}{20~(10)}~ &  \mc{1}{b}{30~(30)}~\\
    \midrule    
    CNN & 0.6030$(\pm0.05)$ & 0.6762$(\pm0.02)$ & 0.6915$(\pm0.02)$ \\
    Pretrain(Dec) & 0.6088$(\pm0.02)$\textsuperscript{\textcolor{black}{$\diamond$}} & 0.6252$(\pm0.03)$\textsuperscript{\textcolor{black}{$\diamond$}} & 0.6439$(\pm0.05)$\textsuperscript{\textcolor{black}{$\diamond$}}\\
    Pretrain(CNN) & 0.6615$(\pm0.03)$\textcolor{black}{*} & 0.6779$(\pm0.02)$ & 0.6890$(\pm0.02)$\\
    MSSL($\gamma$=0.7) & 0.5930$(\pm0.04)$\textsuperscript{\textcolor{black}{$\diamond$}} & 0.6326$(\pm0.03)$\textsuperscript{\textcolor{black}{$\diamond$}} & 0.6860$(\pm0.02)$ \\
    MSSL($\gamma$=0.9) & 0.6189$(\pm0.03)$ & 0.6163$(\pm0.03)$\textsuperscript{\textcolor{black}{$\diamond$}} & 0.6906$(\pm0.02)$ \\
    MSSL(alter) & 0.6509$(\pm0.03)$ & 0.6646$(\pm0.03)$\textsuperscript{\textcolor{black}{$\diamond$}} & 0.6880$(\pm0.02)$\\
    \midrule    
    MASSL($\gamma$=0.7) & 0.6074$(\pm0.03)$\textsuperscript{\textcolor{black}{$\diamond$}}  & 0.6869$(\pm0.03)$ & 0.6900$(\pm0.02)$ \\
    MASSL($\gamma$=0.9) & 0.6654$(\pm0.03)$\textcolor{black}{*} & 0.6925$(\pm0.02)$ & 0.6806$(\pm0.03)$ \\
    \textbf{MASSL(alter)} & \textbf{0.6670}$(\pm0.03)$\textcolor{black}{*}& \textbf{0.7111}$(\pm0.02)$ & \textbf{0.7204}$(\pm0.02)$ \\
    \bottomrule
  \end{tabular}
\end{table}

\begin{table}[h]
  \setlength{\tabcolsep}{4pt}
  \caption{\textbf{Discriminative power of the encoded features.} Using the trained models of all 5 folds on BRATS data, with 50 labeled/70 unlabeled data splits. 5 training/testing data are randomly chosen from the testing sets and used for all models because of the size limitation of the earlier feature maps. The experiment is repeated 5 times with different random data and the mean $R^2$ score (variance) between 5 experiments averaged over all 5-fold models is reported. Note that results can only be compared within columns because the ground truth and dimensionality change between levels.}
  \label{Reg}
  \centering
 \begin{tabular}{lccccc}
    \toprule
    \#Level & \mc{1}{b}{1}~ & \mc{1}{b}{2}~ & \mc{1}{b}{3}~ & \mc{1}{b}{4}~ & \mc{1}{b}{5}~\\
    \midrule    
    CNN & 0.301(.01) & \textbf{0.527}(.01) & 0.496(.01) & 0.422(.01) & 0.486(.04) \\
    MSSL(alter) & \textbf{0.344}(.02) & 0.515(.01) & \textbf{0.524}(.01) & 0.476(.02) & 0.471(.03) \\
    MASSL(alter) & 0.340(.02) & 0.508(.01) & 0.501(.01) & \textbf{0.478}(.01) & \textbf{0.535}(.03) \\
    \bottomrule
  \end{tabular}
\end{table}

\section{Discussion and Conclusion}

In this paper, we propose a new semi-supervised learning method called MASSL that combines a segmentation task and a reconstruction task through an attention mechanism in a multi-task learning network. The proposed method is evaluated on two applications. For both applications, MASSL using part of the labeled images outperforms the fully-supervised CNN baseline using the same number of labeled images, pretraining+finetuning methods, and the proposed approach without attention (MSSL). When using the segmentation and reconstruction loss for all images, MASSL also improves over baseline CNN, although this difference was only statistically significant for the BRATS data. This is mainly due to the sparse distribution of foreground in WMH data, which makes our attention maps less effective.

The improvement of our method mainly comes from the attention mechanism, which introduces the segmentation task into the reconstruction task and links them better than before. The mechanism can be easily integrated into any CNN architecture and generalized to multi-class segmentation. Compared with joint training, alternating training is a practical strategy that allows task-dependent variations in the learning rate and does not require  fine-tuning $\gamma$, although one still needs to choose proper initial learning rates. Alternating training is not guaranteed to be stable because the encoder parameters change discontinuously between the two tasks. During experiments, we found that training was sufficiently stable when choosing a smaller initial learning rate for reconstruction than segmentation, and in most cases, the performance of the alternating optimization was much better than that of joint optimization.

When comparing different multi-task learning strategies, we made some simplifications. For the pretraining method, unlike Sedai et al. \cite{Sedai}, we use a regular autoencoder rather than a variational autoencoder (VAE) in this paper. We think our SSL method could also work well with VAE and perhaps fuse the two tasks even better. In the regression analysis we use a simple regression model that could only show the linear discriminative power of the features. It would be interesting to use a more complicated non-linear model to show the non-linear discriminative power, too. Since we use only one MRI sequence and a subset of scans, our performance on BraTS18 and WMH17 are lower than the state of the art. The best Dice performances of BraTS18 (whole tumor) and WMH17 on testing sets are 0.8839 \cite{Myronenko} and 0.80 \cite{li2018fully} respectively, and first work also uses variational autoencoder to provide more regularization effect similar to the Ladder network \cite{Rasmus} and our MSSL method.

In conclusion, MASSL is a promising segmentation framework for simple and efficient multi-task learning that can achieve strong improvements in semi-supervised as well as in fully supervised settings.

\subsubsection{Acknowledgements}
This research is supported by the China Scholarship Council (File No.201706170040). We gratefully acknowledge the support of the computational resources provided by SURFsara services and Cartesius.

\end{document}